\documentclass[Afour,sageh,times]{sagej}

\usepackage{moreverb,url}

\usepackage{siunitx}

\usepackage{todonotes}

\usepackage[colorlinks,bookmarksopen,bookmarksnumbered,citecolor=red,urlcolor=red]{hyperref}

\newcommand\BibTeX{{\rmfamily B\kern-.05em \textsc{i\kern-.025em b}\kern-.08em
T\kern-.1667em\lower.7ex\hbox{E}\kern-.125emX}}

\def\volumeyear{2016}

\usepackage{dirtree}

\usepackage{xcolor}
\newcommand{\numberObjects}{14~}
\newcommand{\numberInteractions}{358~}

\usepackage{rotating}
\usepackage{tabularx}
\usepackage{ragged2e}
\usepackage{lipsum}
\usepackage{lscape} 
\usepackage{textcomp}

\usepackage{tikz}
\usepackage{array}
\newcolumntype{P}[1]{>{\centering\arraybackslash}p{#1}}
\newcolumntype{M}[1]{>{\centering\arraybackslash}m{#1}}

\usepackage{listings}

\begin{document}

\runninghead{Mart\'{i}n-Mart\'{i}n et al.}

\title{The RBO Dataset of Articulated Objects and Interactions}

\author{Roberto Mart\'{i}n-Mart\'{i}n*, Clemens Eppner* and Oliver Brock}

\affiliation{All authors are with the Robotics and Biology Laboratory, Technische
Universit\"{a}t Berlin, Germany. * denotes equal contribution.}

\corrauth{Roberto Mart\'{i}n-Mart\'{i}n, Robotics and Biology Laboratory, Marchstr. 23, 10587 Berlin, Germany}

\email{roberto.martinmartin@tu-berlin.de}

\begin{abstract}
We present a dataset with models of \numberObjects articulated objects commonly found in human environments and with RGB-D video sequences and wrenches recorded of human interactions with them. The 358 interaction sequences total 67~minutes of human manipulation under varying experimental conditions (type of interaction, lighting, perspective, and background). Each interaction with an object is annotated with the ground truth poses of its rigid parts and the kinematic state obtained by a motion capture system. For a subset of 78~sequences (25~minutes), we also measured the interaction wrenches. The object models contain textured three-dimensional triangle meshes of each link and their motion constraints. We provide Python scripts to download and visualize the data. The data is available at \href{https://tu-rbo.github.io/articulated-objects/}{https://tu-rbo.github.io/articulated-objects/} and hosted at \href{https://zenodo.org/record/1036660/}{https://zenodo.org/record/1036660/}.
\end{abstract}


\maketitle

\section{Introduction}

The RBO dataset is a collection of 358~RGB-D video sequences (67~minutes) of humans manipulating \numberObjects articulated objects under varying exeperimental conditions (type of interaction, lighting, perspective, and background). All sequences are annotated with ground truth of the poses of the rigid parts and the kinematic state of the articulated object (joint states) obtained with a motion capture system. We also provide kinematic models of these objects including three-dimensional textured shape models. For 78~sequences (25~minutes) the interaction wrenches during the manipulation are also recorded.

We present the first dataset with \emph{articulated} objects. All similar datasets contain \emph{single rigid-body} objects that move or are being manipulated. There are two datasets that could be considered close to ours. The first one~\citep{GarciaCifuentes.RAL} is a dataset that was released together with a method for robot arm tracking. This dataset contains images and joint encoder values of a moving robot arm and its kinematic and shape model. In contrast, our dataset is targeted to the study of interactions with everyday human objects: it contains models of multiple ubiquitous articulated objects and sequences of interactions in varying environmental conditions. The second dataset~\citep{michel2015pose} provides models of four everyday articulated objects and ground truth of the static pose of each link and the changing pose of camera during the video sequence. The sequences of this dataset do not contain any interaction or manipulation of the objects, only camera motion. Therefore, neither of these datasets can be used to study interactions and to evaluate methods for perceiving them.

\begin{figure}[t!]
    \centering
    \includegraphics[width=\linewidth]{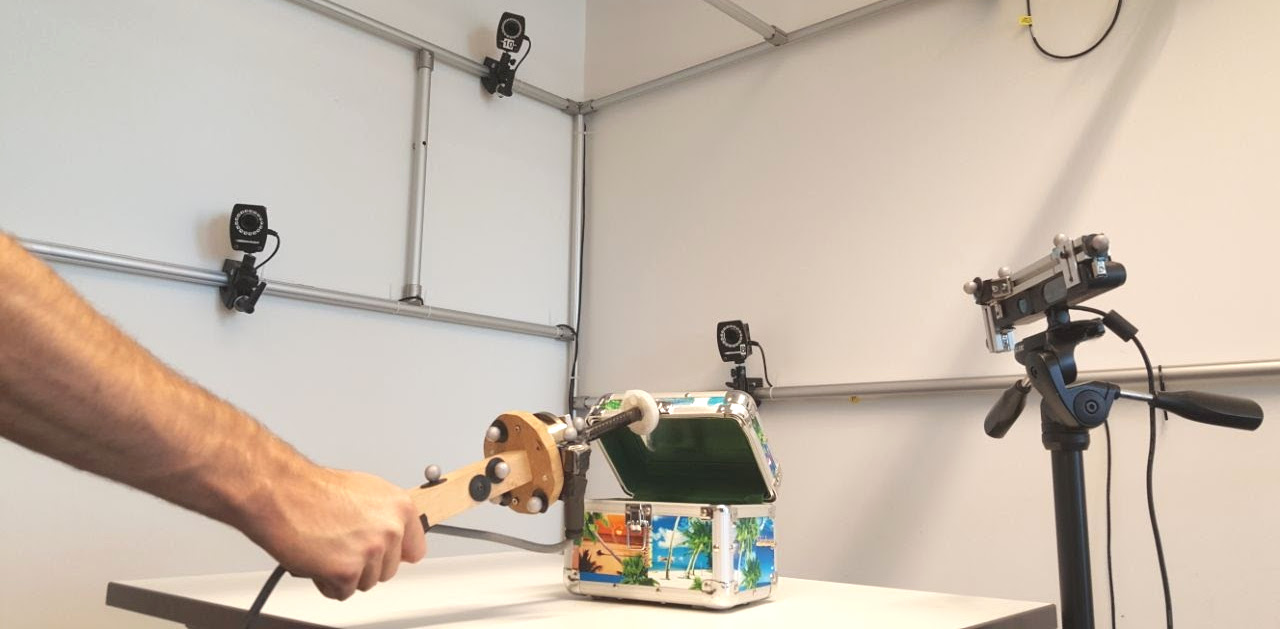}
    \includegraphics[width=.5\linewidth]{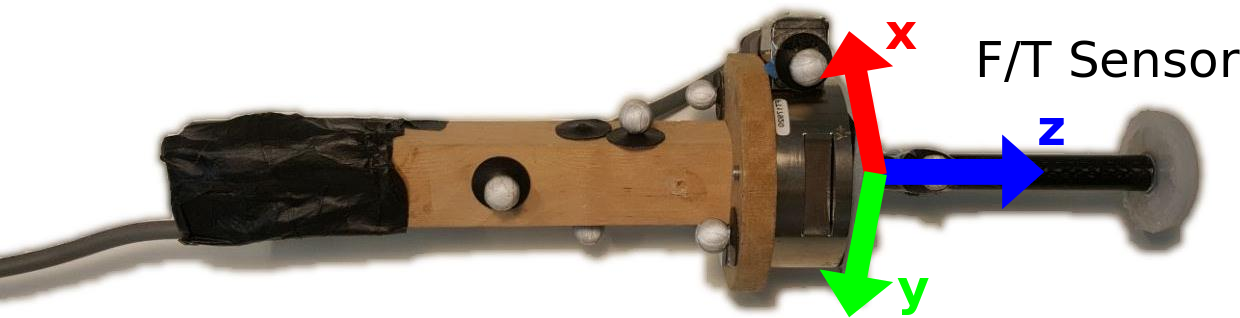}%
    \hfill\includegraphics[width=.4\linewidth]{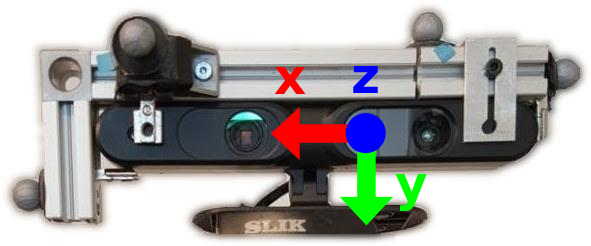}
    \caption{Our sensor setup for recording interactions with articulated objects; Top: interaction in the motion capture volume; Bottom left: tool with F/T sensor for recording interaction wrenches, motion capture markers, and measurements reference frame; Bottom right: Asus RGB-D sensor with motion capture markers and measurements reference frame}
    \label{fig:systemsetup}
\end{figure}

Our dataset will help evaluate algorithms for tracking articulated objects (e.g.~\cite{schmidt2014dart}) and building models of articulated objects (e.g.~\cite{martin-martin_integrated_2016}).
Apart from benchmarking, the dataset can also be used to develop data-driven algorithms by exploiting the provided models to generating virtual visual data. While there is a vast amount of three-dimensional models and datasets of objects~\citep{Kit2012,calli2015ycb}, very few of them include and describe articulated mechanisms.

\section{Sensor Setup}
\label{s:sensor_setup}

We use the following sensors to record human interactions with articulated objects~(see Fig.~\ref{fig:systemsetup}):
\begin{itemize}
\item RGB-D camera Asus Xtion Pro Live, 640$\times$480~pixels, \SI{30}{\Hz}, pointed at the object.
\item Motion capture system by \cite{mocap}, capture volume: $1.5~m^3$, including 18~Osprey cameras, providing 3-D positions of fiducial markers at \SI{100}{\Hz}.
\item F/T sensor ATI FTN-Gamma DAQ/Net, calibration SI-130-10, force/torque resolution: $F_x = F_y = \frac{1}{40}N$, $F_z = \frac{1}{20}N$, \;$T_x = T_y = T_z = \frac{1}{800}Nm$, recorded at~\SI{100}{\Hz}.
\end{itemize}


We acquired separately 3-D~triangle meshes of each articulated object part, using the following sensors and methods:
\begin{itemize}
\item Structured light scanning system by \cite{shaperec2}, SLS-3 HD, scan size: \SI{60}{}-\SI{500}{\mm}, resolution: down to \SI{0.05}{\mm}.
\item Reconstruction software \cite{shaperec}, used with photos taken with Casio Exilim EX-FC100, 9~MP.
\end{itemize}

\section{Data Structure and Usage}

\begin{figure}[th!]
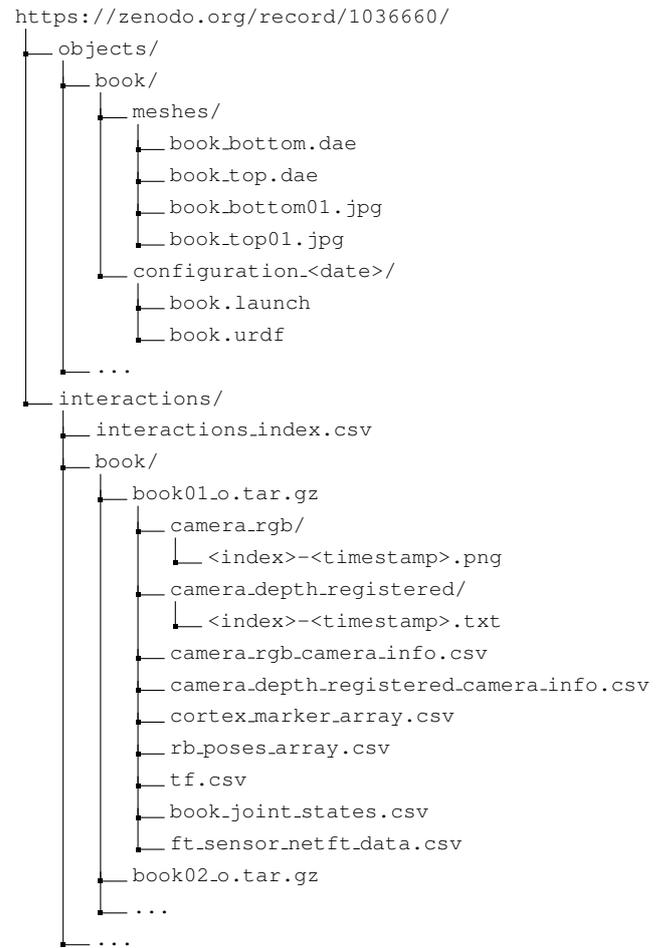

\footnotesize
\dirtree{%
.1 https://zenodo.org/record/1036660/.
.2 objects/.
.3 book/.
.4 meshes/.
.5 book\_bottom.dae.
.5 book\_top.dae.
.5 book\_bottom01.jpg.
.5 book\_top01.jpg.
.4 configuration\_<date>/.
.5 book.launch.
.5 book.urdf.
.3 \ldots{}.
.2 interactions/.
.3 interactions\_index.csv.
.3 book/.
.4 book01\_o.tar.gz.
.5 camera\_rgb/.
.6 <index>-<timestamp>.png.
.5 camera\_depth\_registered/.
.6 <index>-<timestamp>.txt.
.5 camera\_rgb\_camera\_info.csv.
.5 camera\_depth\_registered\_camera\_info.csv.
.5 cortex\_marker\_array.csv.
.5 rb\_poses\_array.csv.
.5 tf.csv.
.5 book\_joint\_states.csv.
.5 ft\_sensor\_netft\_data.csv.
.4 book02\_o.tar.gz.
.4 \ldots{}.
.3 \ldots{}.
}
\caption{The dataset is structured by objects and interactions. Please see text for details.}
\label{fig:dirtree}
\end{figure}

The RBO dataset is available at \href{https://tu-rbo.github.io/articulated-objects/}{https://tu-rbo.github.io/articulated-objects/} and hosted at \href{https://zenodo.org/record/1036660/}{https://zenodo.org/record/1036660/}.
It is composed of two parts~(see~Fig.~\ref{fig:dirtree}): a first part with descriptions of~\numberObjects~articulated objects and the main part containing~\numberInteractions~human interactions with these objects.
We also provide Python scripts to facilitate downloading, visualizing and working with the data.

\subsection{Models of Articulated Object (\texttt{\small objects/})}

The RBO dataset contains~\numberObjects models of articulated mechanisms that are commonly found in human environments. 
Table~\ref{tab:wide-item-tbl} depicts these objects and their kinematic structure.
Each object model (\texttt{\footnotesize<object\_id>/}) consists of:
\begin{itemize}
\item \textbf{Link geometries} (\texttt{\footnotesize meshes/}): We describe the shape of a link as a three-dimensional triangle textured mesh in the~\cite{isocollada} format~(\texttt{\footnotesize<part\_name>.dae}). We provide the associated texture as JPEG images~(\texttt{\footnotesize <part\_name><index>.jpg}).
\item \textbf{Kinematic structure} (\texttt{\footnotesize configuration\_<date>/}): We define the relation of links and joints with the widely-used \emph{Unified Robot Description Format}~\cite{urdf} (\texttt{\footnotesize <object\_id>.urdf}), an XML file format to describe all elements of articulated objects with chain or tree structure. 
The objects of the database possess one degree-of-freedom (DoF) joints that can be either prismatic or revolute joints.
The base link is the origin of the kinematic tree or chain.
It is either rigidly connected to the environment (represented with a \emph{static} joint with zero DoF) or completely unconstrained (represented with a floating joint with six DoF).
The reference coordinate frame of a link corresponds to a marker set of the motion capture system (see Section \emph{Data Acquisition}).
We define the joint parameters and link meshes with respect to these coordinate frames.
Since marker set locations can vary between recording sessions, we provide a separate kinematic structure description for each session indicated by the \texttt{\footnotesize \_<date>} suffix in the folder name.
\end{itemize}

\begin{table}
    \centering
\caption{The dataset contains 14~articulated objects. Letters in the last column represent (S)tatic, (F)loating, (R)evolute and (P)rismatic joints.}
    \label{tab:wide-item-tbl}
\begin{tabular}{M{0.13\textwidth}  M{0.13\textwidth}  M{0.13\textwidth}}
Object ID & Picture & Actuated Joints \\
\hline
\centering
globe & \vspace{2pt}\includegraphics[height=1.5cm]{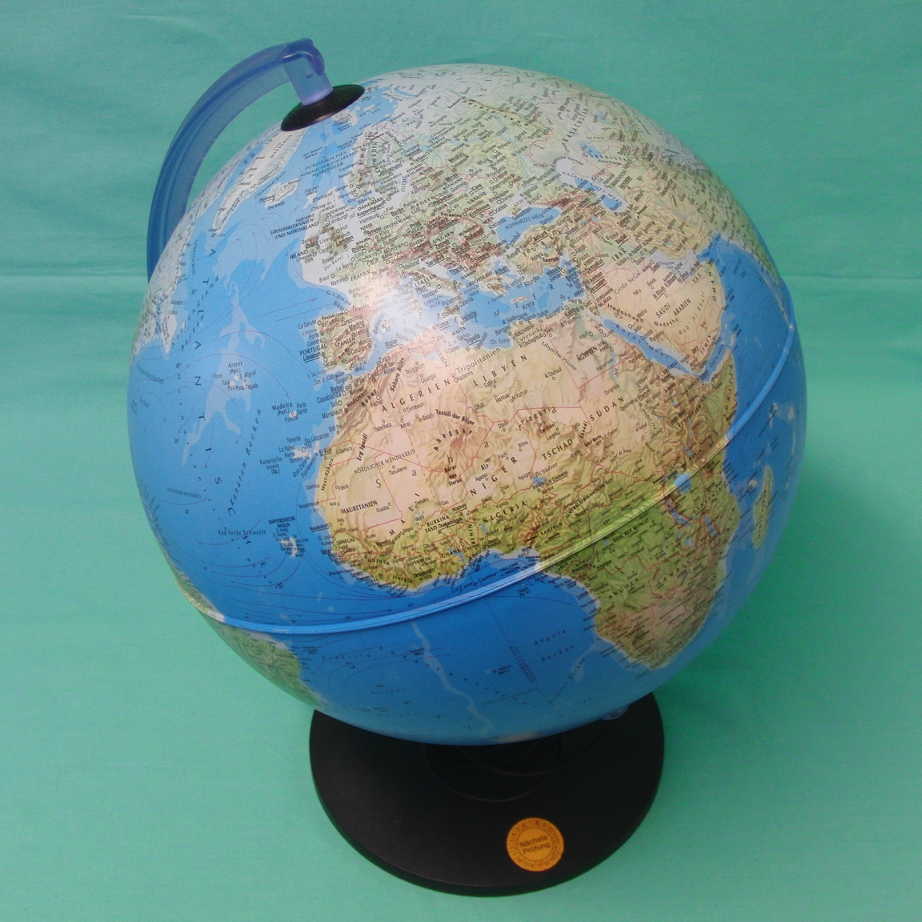} & 
\begin{tikzpicture}[node distance=0.5cm]
{\node (F) at (0,0) {F};
\node (R) at (1,0) {R};
\draw [-] (F) -- (R);
}
\end{tikzpicture} \\ 
\centering
laptop & \includegraphics[height=1.5cm]{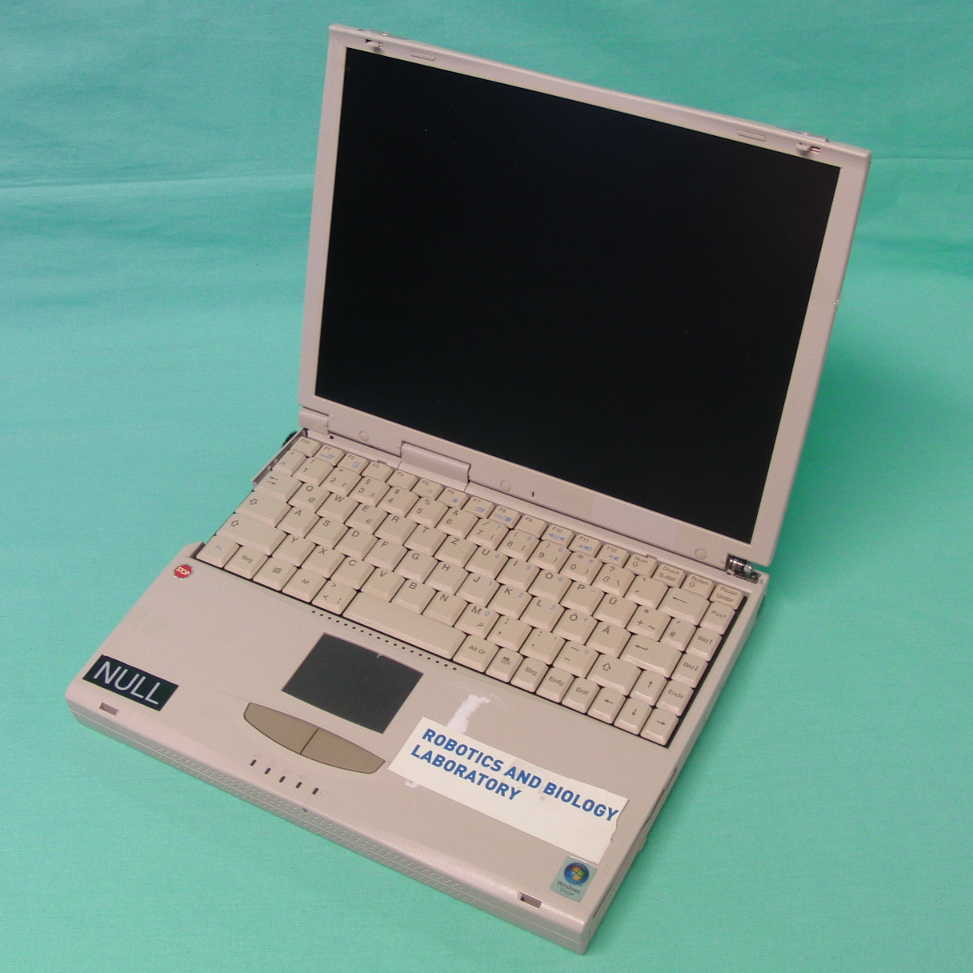} & \begin{tikzpicture}[node distance=0.5cm]
{\node (F) at (0,0) {F};
\node (R) at (1,0) {R};
\draw [-] (F) -- (R);
}
\end{tikzpicture} \\
\centering
ikea & \includegraphics[height=1.5cm]{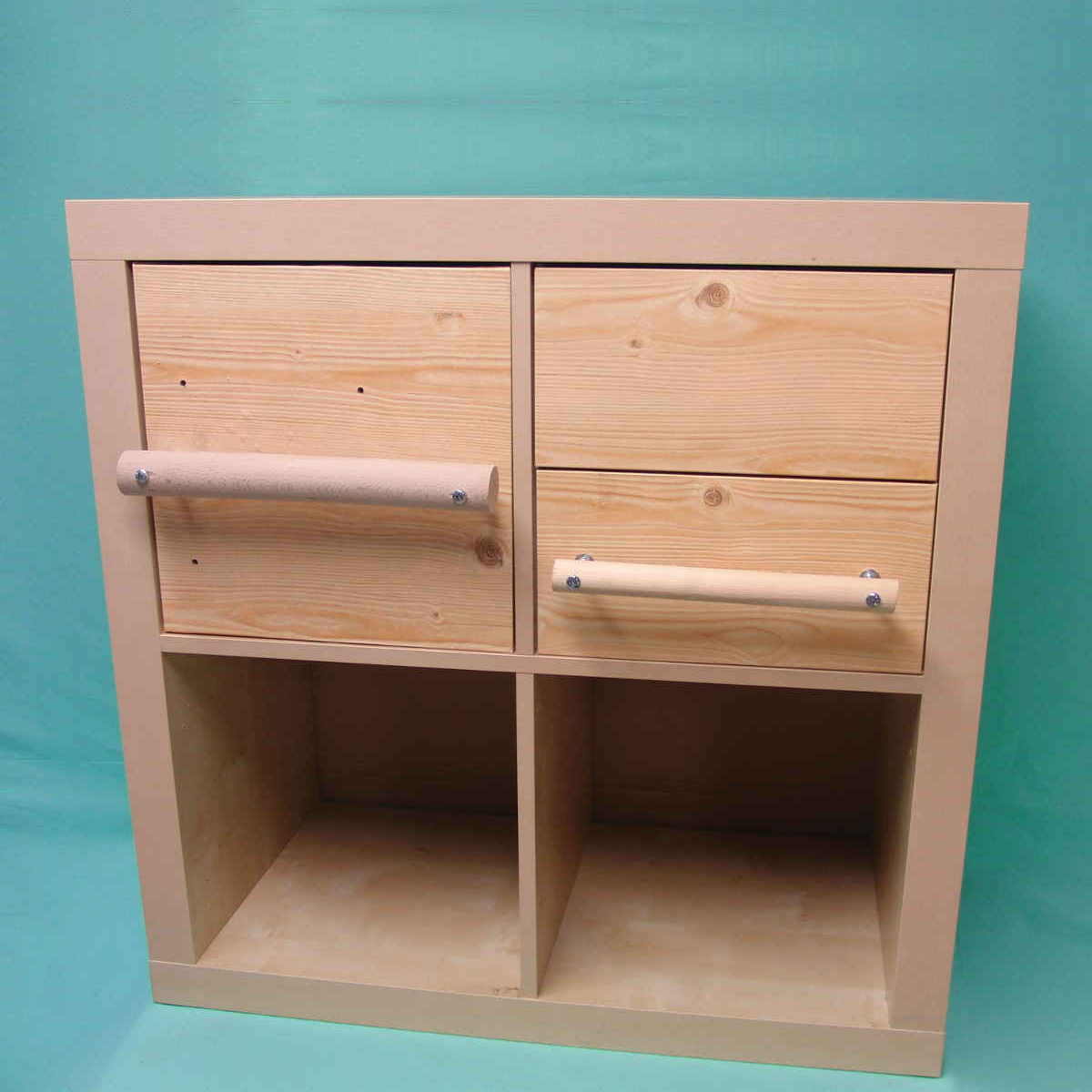} & \begin{tikzpicture}[node distance=0.05cm]
{\node (S) at (0,1) {S};
\node (R) at (1,0.5) {P};
\node (R2) at (1,1.5) {R};
\draw [-] (S) -- (R) (S) -- (R2);
}
\end{tikzpicture} \\
\centering
foldingrule & \includegraphics[height=1.5cm]{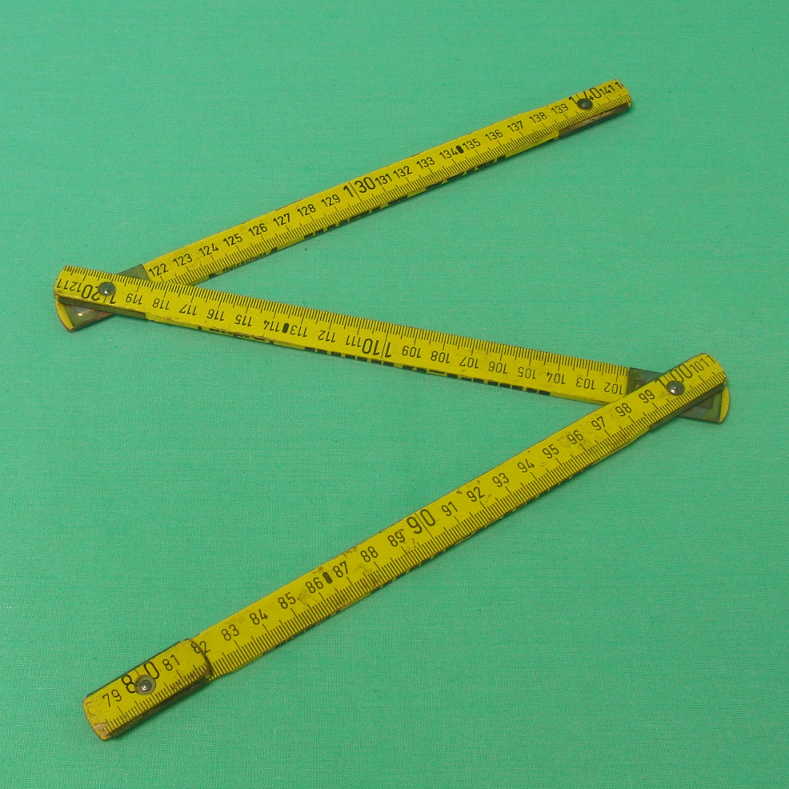} & \begin{tikzpicture}[node distance=0.05cm]
{\node (F) at (0,0) {F};
\node (R) at (1,0) {R};
\node (R2) at (2,0) {R};
\draw [-] (F) -- (R) -- (R2);
}
\end{tikzpicture} \\
\centering
book & \includegraphics[height=1.5cm]{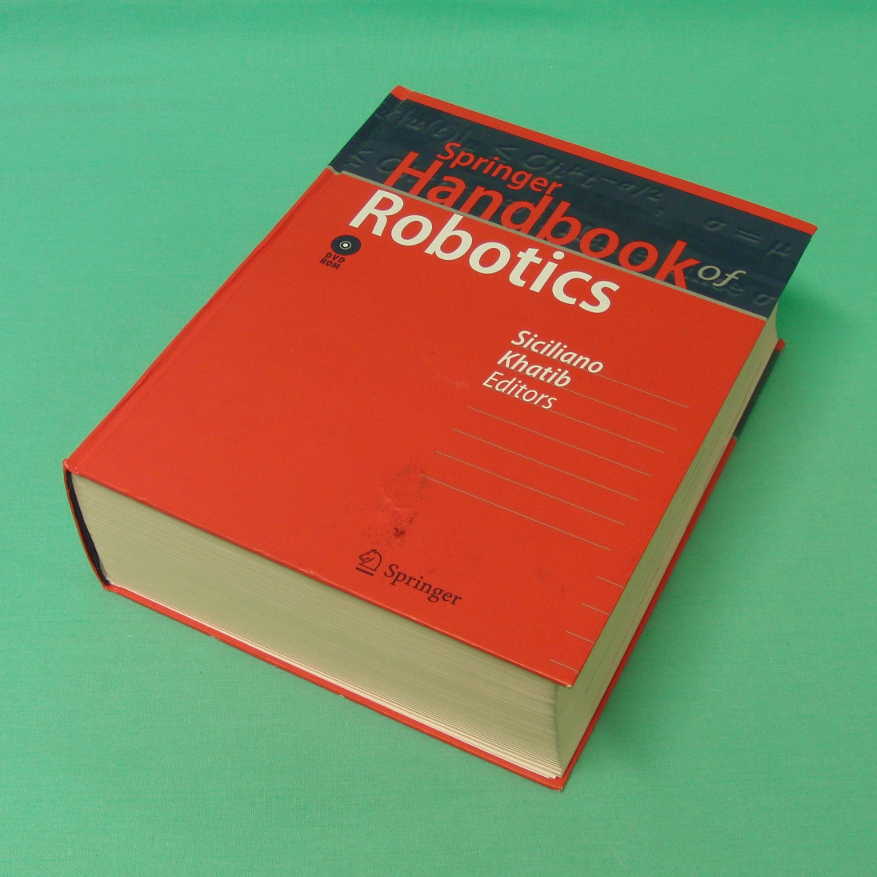} &\begin{tikzpicture}[node distance=0.5cm]
{\node (F) at (0,0) {F};
\node (R) at (1,0) {R};
\draw [-] (F) -- (R);
}
\end{tikzpicture} \\
\centering
treasurebox &\includegraphics[height=1.5cm]{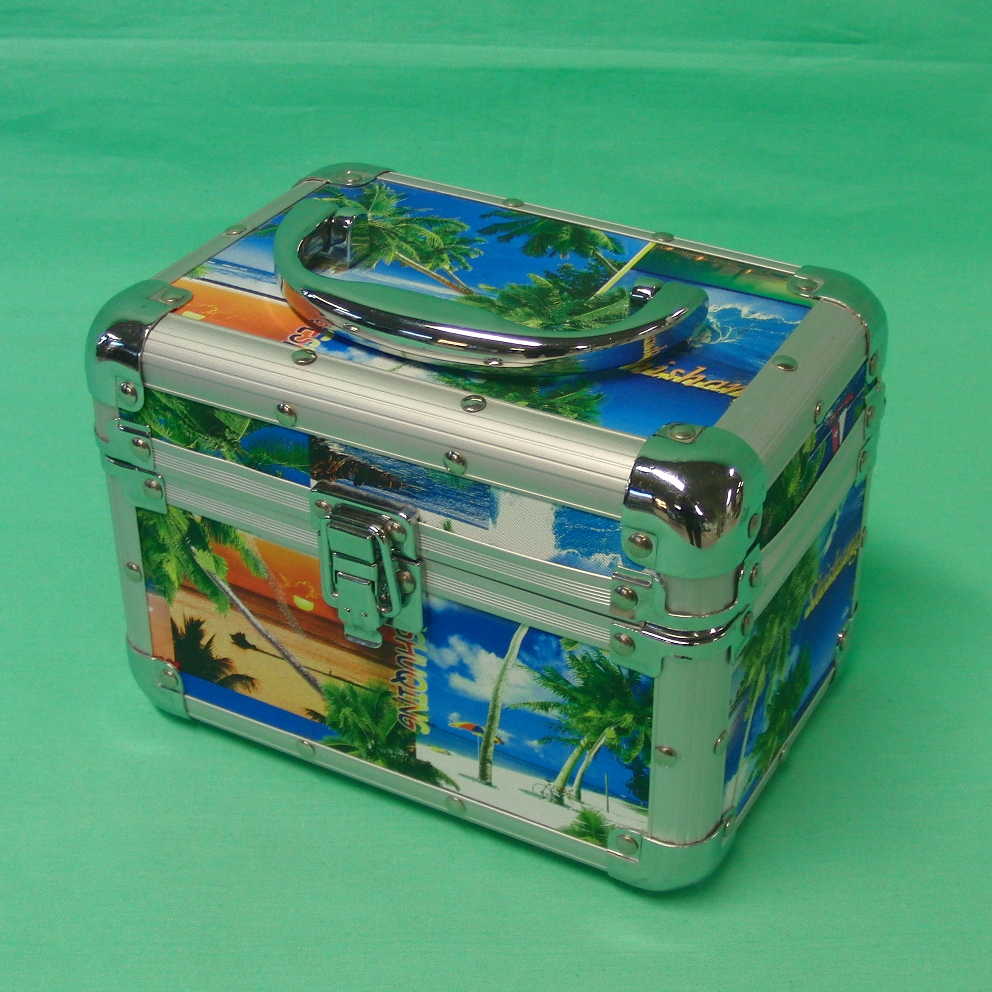} & \begin{tikzpicture}[node distance=0.5cm]
{\node (F) at (0,0) {F};
\node (R) at (1,0) {R};
\draw [-] (F) -- (R);
}
\end{tikzpicture} \\
\centering
tripod & \includegraphics[height=1.5cm]{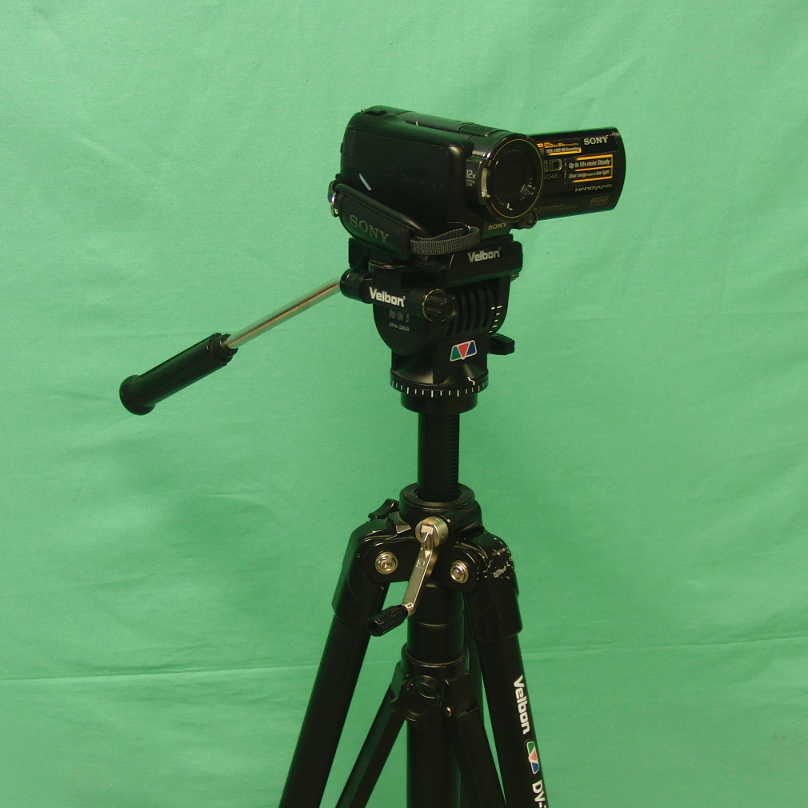} & \begin{tikzpicture}[node distance=0.5cm]
{\node (F) at (0,0) {F};
\node (R) at (1,0) {P};
\node (R2) at (2,0) {R};
\draw [-] (F) -- (R) -- (R2);
}
\end{tikzpicture} \\
\centering
clamp & \includegraphics[height=1.5cm]{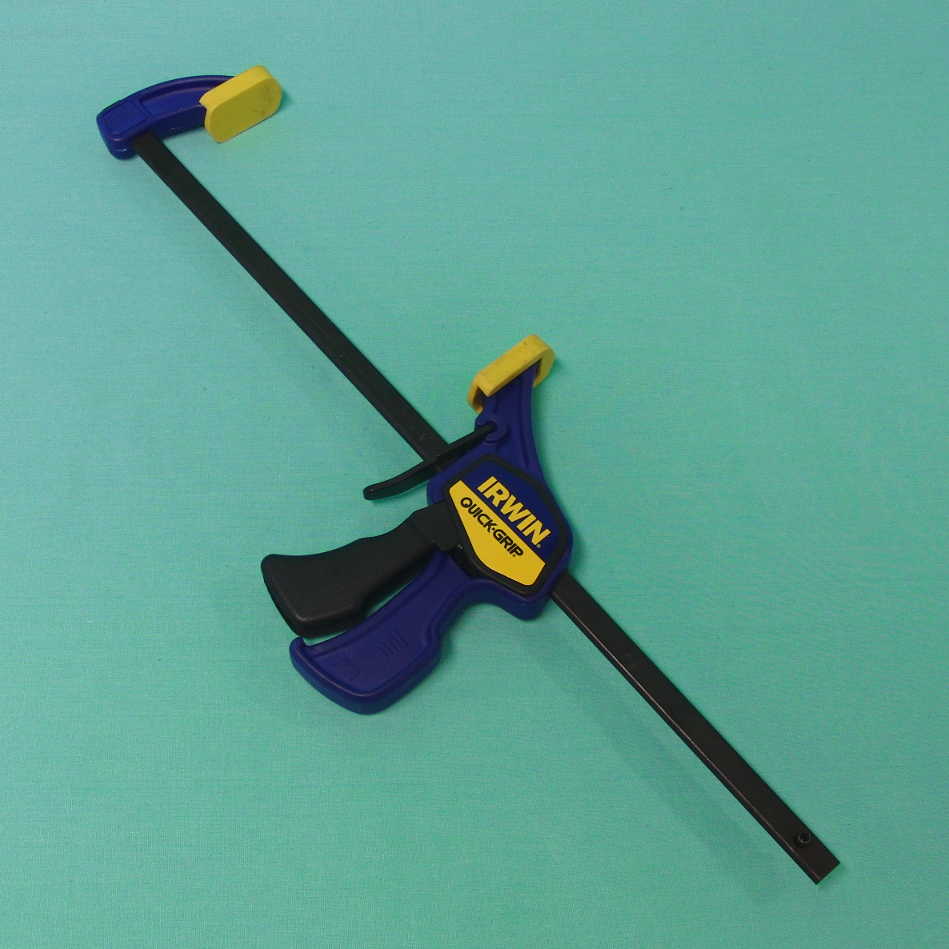} & \begin{tikzpicture}[node distance=0.5cm]
{\node (F) at (0,0) {F};
\node (P) at (1,0) {P};
\draw [-] (F) -- (P);
}
\end{tikzpicture} \\
\centering
pliers & \includegraphics[height=1.5cm]{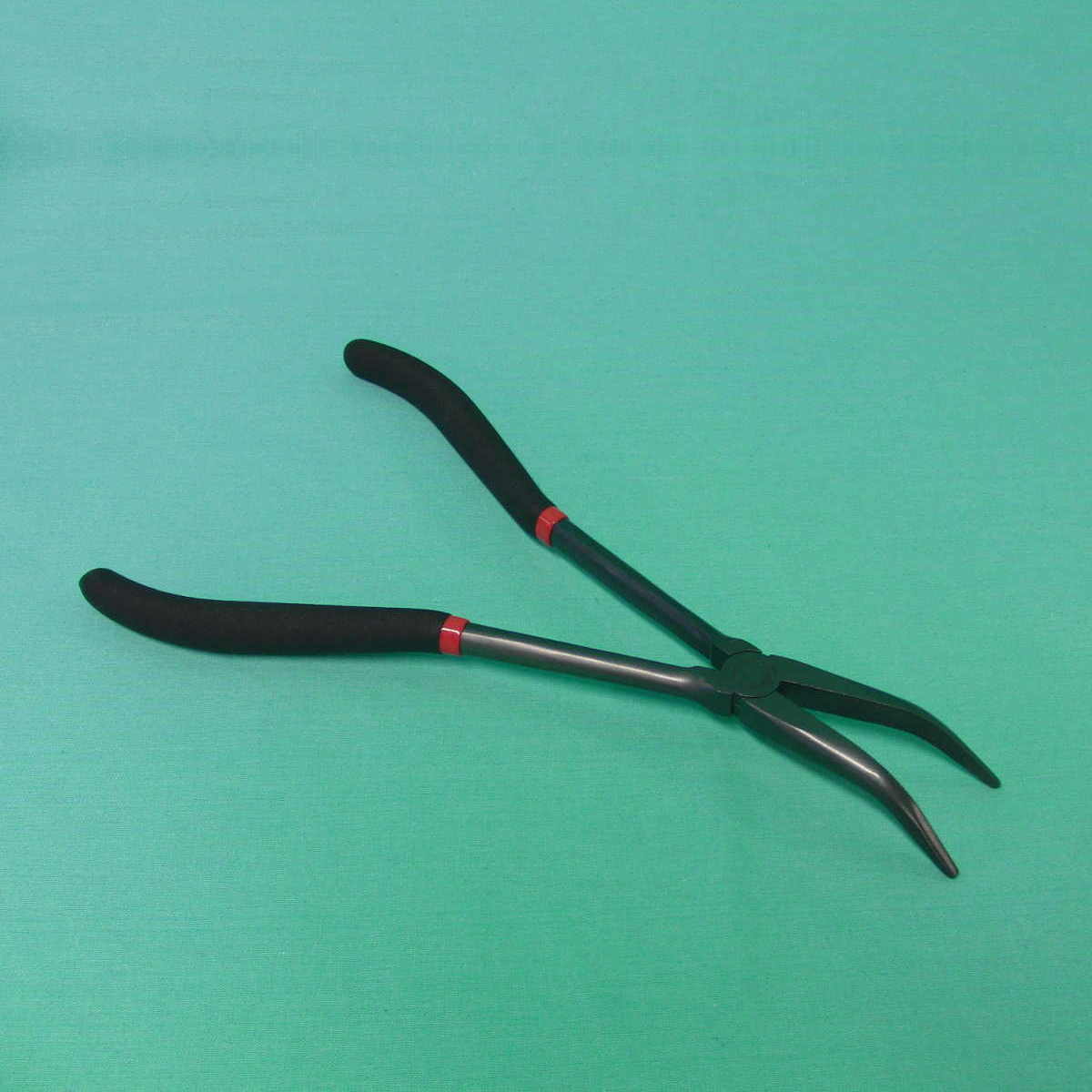} & \begin{tikzpicture}[node distance=0.5cm]
{\node (F) at (0,0) {F};
\node (R) at (1,0) {R};
\draw [-] (F) -- (R);
}
\end{tikzpicture} \\
\centering
cardboardbox & \includegraphics[height=1.5cm]{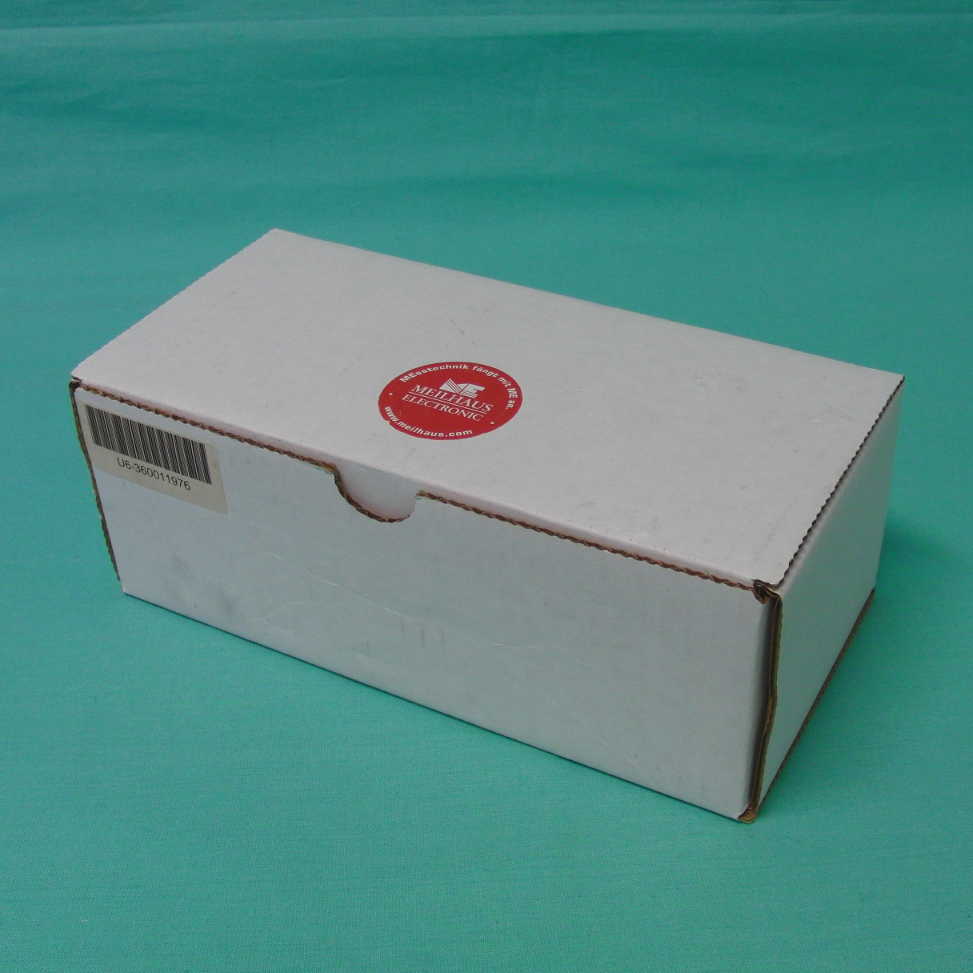} & \begin{tikzpicture}[node distance=0.5cm]
{\node (F) at (0,0) {F};
\node (R) at (1,0) {R};
\draw [-] (F) -- (R);
}
\end{tikzpicture} \\
\centering
rubikscube & \includegraphics[height=1.5cm]{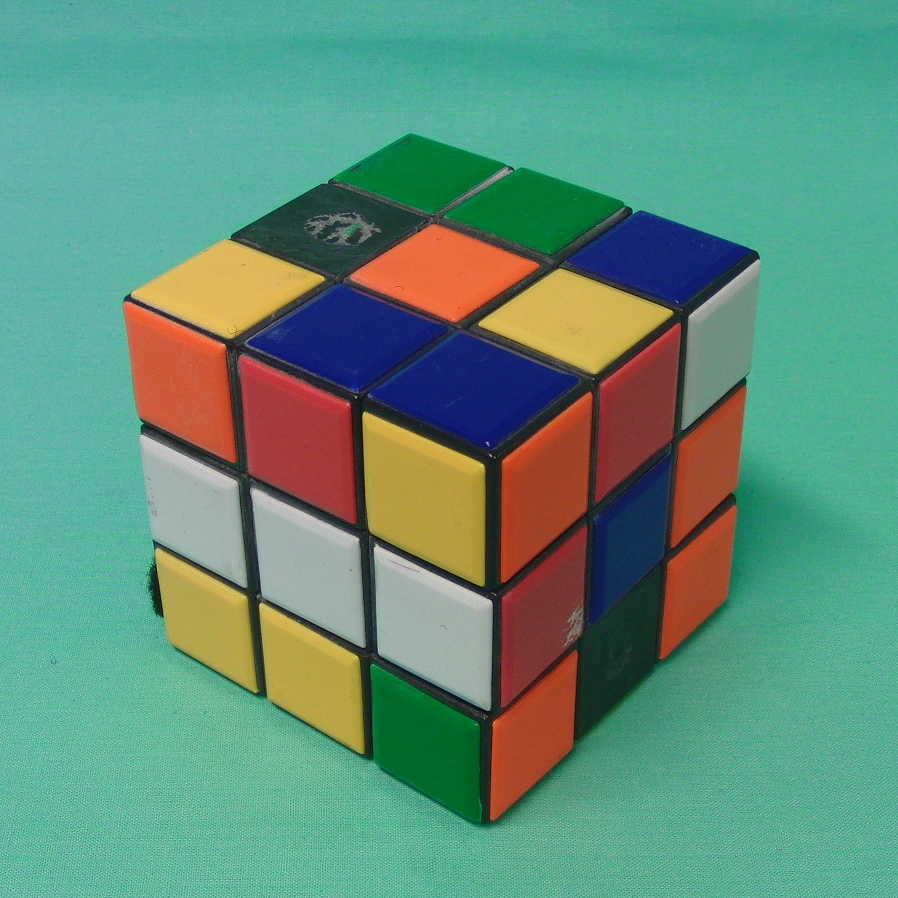} & \begin{tikzpicture}[node distance=0.5cm]
{\node (F) at (0,0) {F};
\node (R) at (1,0) {R};
\draw [-] (F) -- (R);
}
\end{tikzpicture} \\
\centering
microwave & \includegraphics[height=1.5cm]{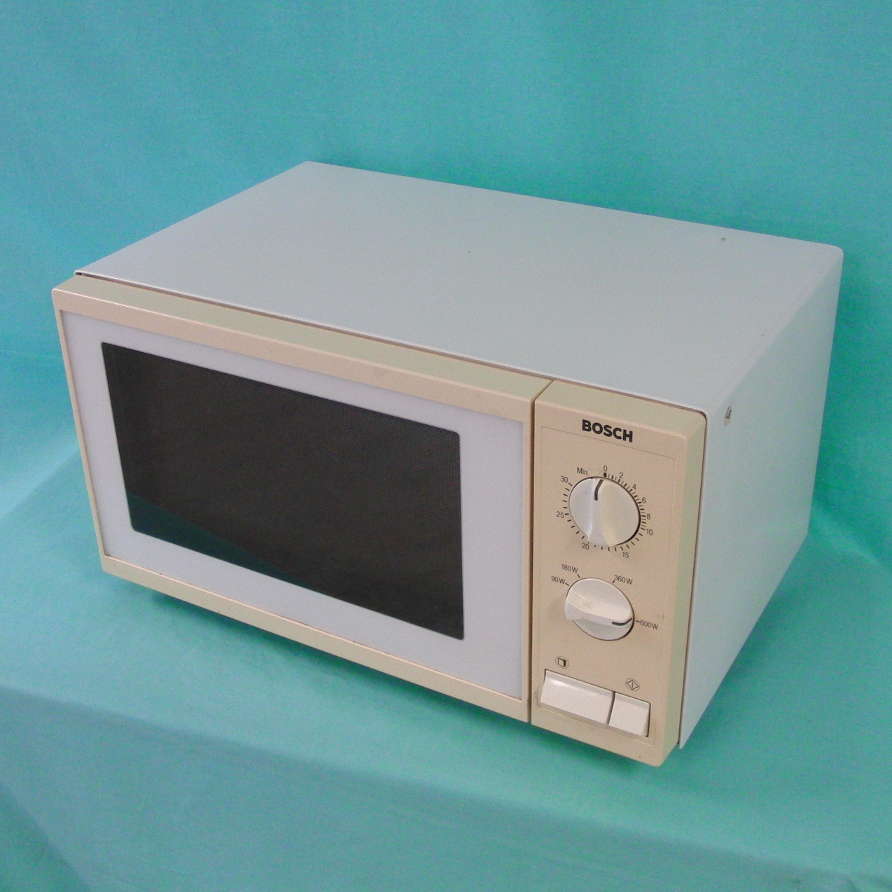} & \begin{tikzpicture}[node distance=0.5cm]
{\node (S) at (0,0) {S};
\node (R) at (1,0) {R};
\draw [-] (S) -- (R);
}
\end{tikzpicture} \\
\centering
ikeasmall & \includegraphics[height=1.5cm]{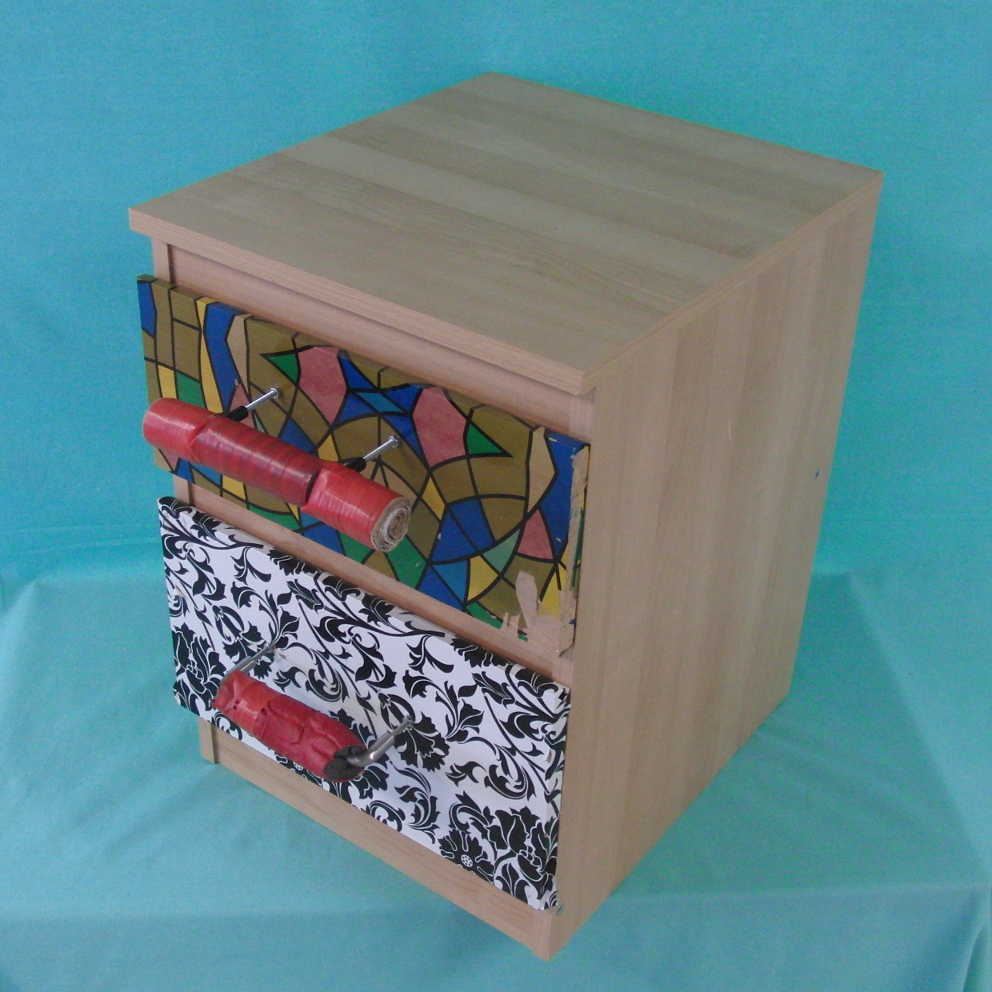} & \begin{tikzpicture}[node distance=0.05cm]
{\node (S) at (0,1) {S};
\node (P) at (1,0.5) {P};
\node (P2) at (1,1.5) {P};
\draw [-] (S) -- (P) (S) -- (P2);
}
\end{tikzpicture} \\
\centering
cabinet & \includegraphics[height=1.5cm]{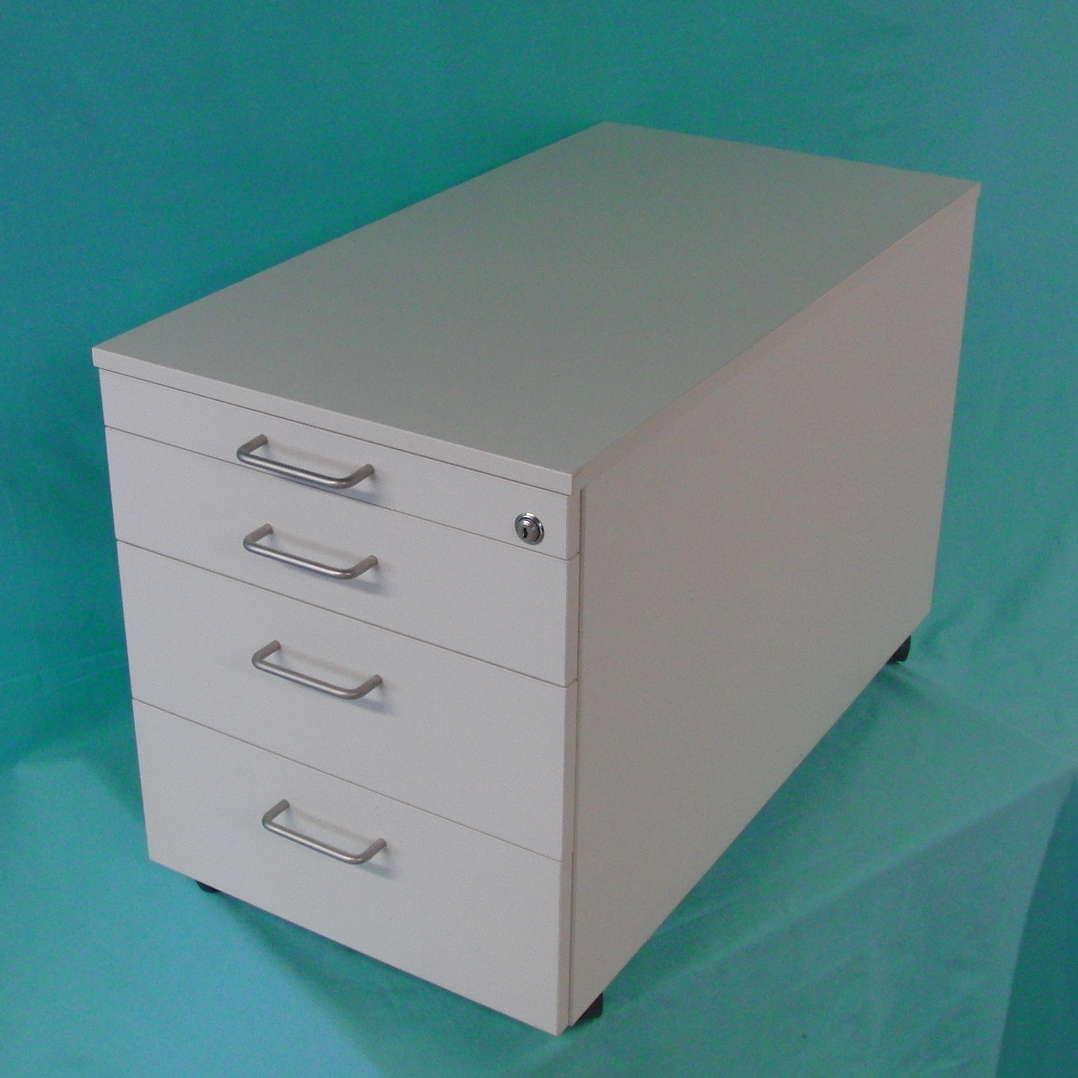} & \begin{tikzpicture}[node distance=0.5cm]
{\node (S) at (0,1) {S};
\node (P) at (1,0.5) {P};
\node (P2) at (1,1.5) {P};
\draw [-] (S) -- (P) (S) -- (P2);
}
\end{tikzpicture} \\
    \hline
\end{tabular}
\end{table}

\subsection{Interactions (\texttt{\small interactions/})}

The RBO dataset contains sensor data of~\numberInteractions human interactions with the \numberObjects modelled objects ($\geq25$~interactions per object).
The sequences last between~\SI{2.7}{\s} and~\SI{69.0}{\s} (median:~\SI{9.15}{\s}).
They differ in lighting conditions, camera perspective and motion, background, clutter, actuation of the mechanisms and human motion~(see Table~\ref{tab:int-props}).
The file \texttt{\footnotesize interactions\_index.csv} contains a list of all interactions and their properties.

The sensor data is organized per object (\texttt{\footnotesize <object\_id>/}) and interaction (\texttt{\footnotesize <object\_id><index>\_o/}). 
Each interaction includes:
\begin{itemize}
    \item \textbf{RGB images}: We store the color images as 8-bit loss-less compressed PNG files (\texttt{\footnotesize camera\_rgb/<index>-<timestamp>.png}).
    \item \textbf{Depth images}: We register the depth images to the RGB camera frame (see Section \emph{Data Acquisition}) and store them as text files containing distances in meters (\texttt{\footnotesize camera\_depth\_registered/<index>-<timestamp>.txt}).
    \item \textbf{Intrinsic camera parameters}: We provide the focal length, center point and distortion parameters of the \emph{Plumb Bob} model~\citep{brown1966decentering} for both, the camera that generates RGB~(\texttt{\footnotesize camera\_rgb\_camera\_info.csv}) and the one that generates depth images~(\texttt{\footnotesize camera\_depth\_registered\_camera\_info.csv}).
    \item \textbf{Extrinsic camera parameters}: We represent the 6-D transformation between the cameras of the RGB-D sensor with a translation vector and a quaternion (\texttt{\footnotesize tf.csv}).
    \item \textbf{Infra-red marker positions}: We include the 3-D locations of all infrared fiducial markers in the scene measured by the motion capture system (\texttt{\footnotesize cortex\_markers\_array.csv}).
    \item \textbf{Rigid body poses}: We store the 6-D poses defined by sets of infrared fiducial markers as position vectors and quaternions. We include the pose of each link of the articulated object, the \mbox{RGB-D} and force/torque (F/T) sensor at \SI{100}{\Hz} (\texttt{\footnotesize rb\_poses\_array.csv}).
    \item \textbf{Joint configurations}: We compute the joint configuration of the articulated object in the scene from the 6-D poses of its links (\texttt{\footnotesize <object>\_joint\_states.csv}).
    \item \textbf{Wrenches}: We provide the forces and torques for the interaction as provided by the F/T sensor (\texttt{\footnotesize ft\_ssensor\_netft\_data.csv}). We include this data in at least five interactions per object.
\end{itemize}

\begin{figure*}[t!]
\centering
\begin{tabular}[0.99\textsize]{lll}
\hspace{-12pt}
\begin{tabular}{c}%
\includegraphics[width=.25\linewidth]{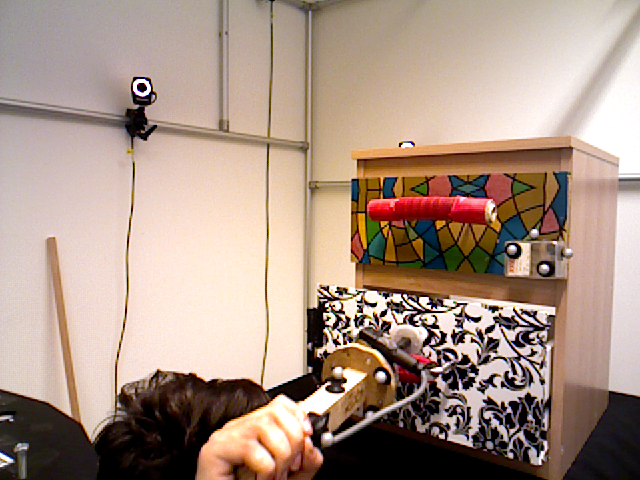}\\
\includegraphics[width=.25\linewidth]{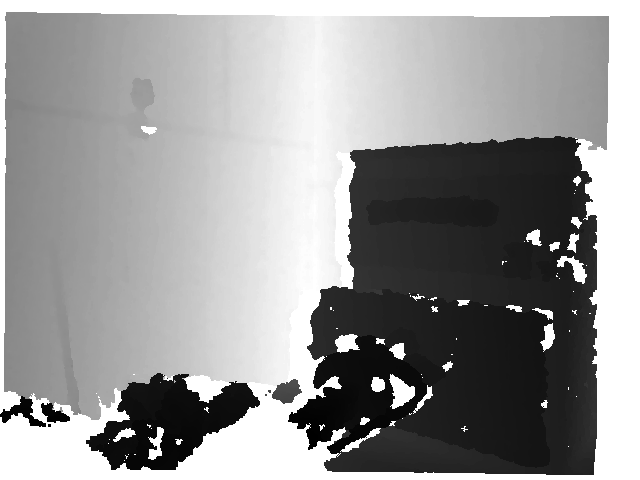} 
\end{tabular}&
\hspace{-20pt}
\begin{tabular}{c}
\includegraphics[width=.47\linewidth]{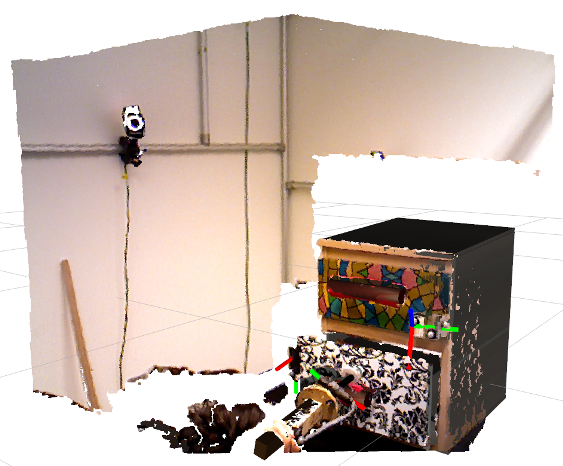}%
\end{tabular}&
\hspace{-20pt}
\begin{tabular}{c}
\includegraphics[width=.25\linewidth]{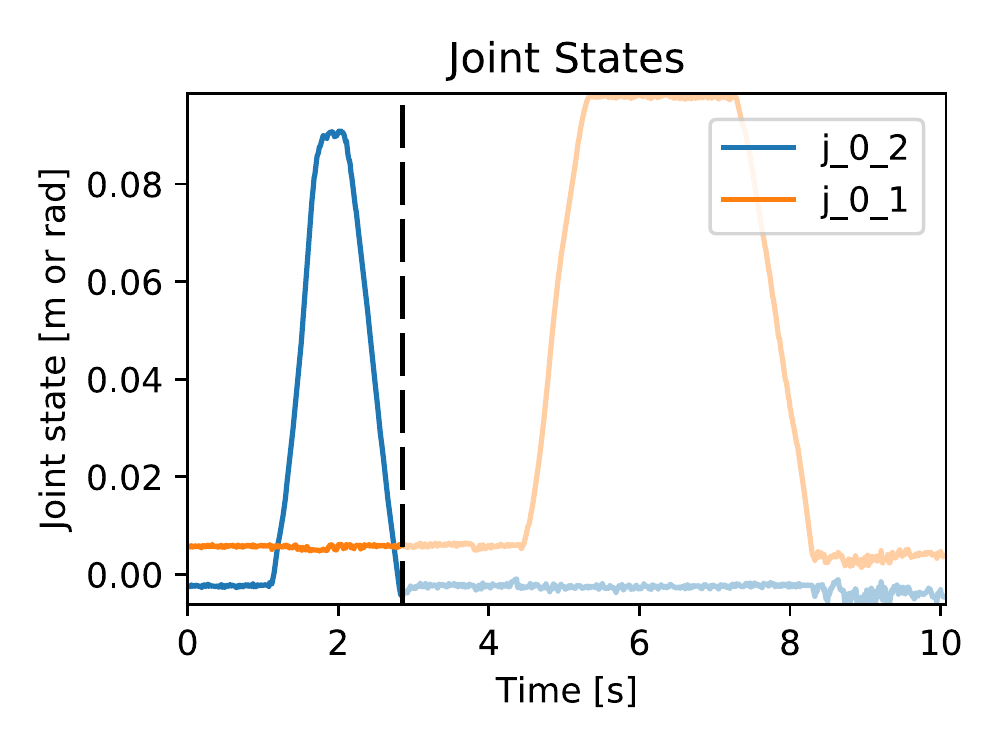}\\
\includegraphics[width=.25\linewidth]{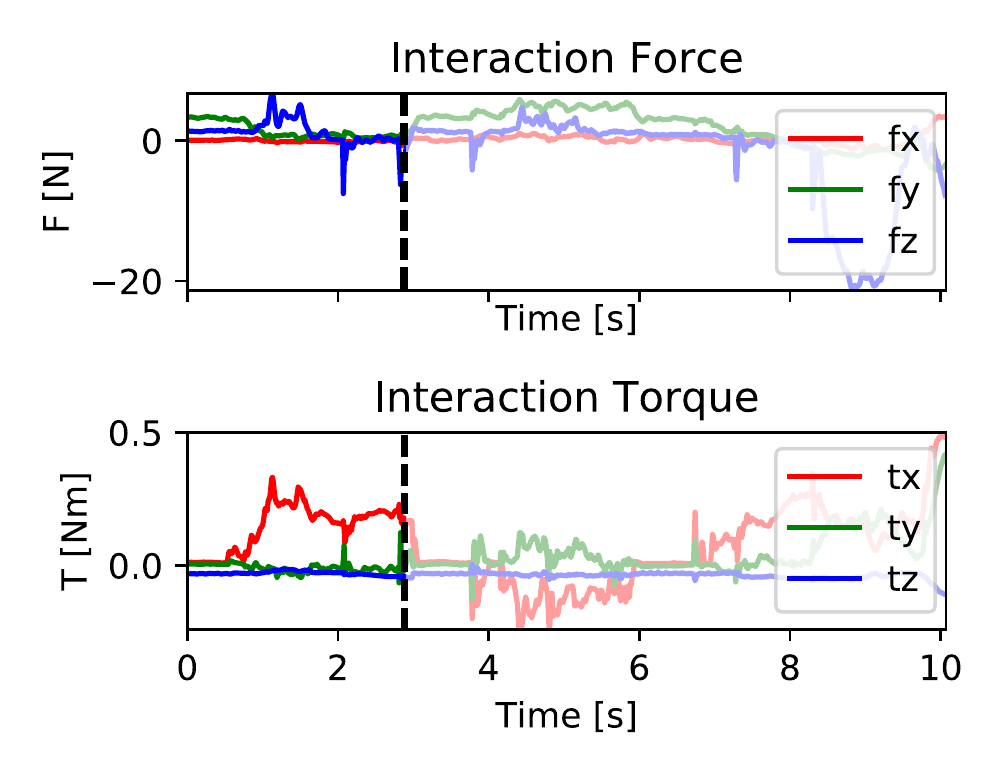}
\end{tabular}
\end{tabular}
    \caption{Visualization of the sensor data for opening a drawer: RGB and depth image (\textit{left}), mesh model, point cloud and coordinate frames of tracked bodies (\textit{middle}), and drawer state and applied wrenches (\textit{right}) with dashed vertical line indicating current time.}
    \label{fig:interactiondata}
\end{figure*}

\begin{table}
    \centering
\caption{Properties of the \numberInteractions interaction recordings}
    \label{tab:int-props}
\begin{tabular}[0.49\textsize]{lll}
Lighting Conditions & \begin{tabular}{l} Artificial
 \\Natural\\Dark\end{tabular} & \begin{tabular}{r}178\\107\\73\end{tabular}\\
\hline
Camera Motion & \begin{tabular}{l} Yes \\No\end{tabular} & \begin{tabular}{r} 100 \\258\end{tabular}\\
\hline
Type of Background & \begin{tabular}{l} Plain \\Textured\\Black\end{tabular}  & \begin{tabular}{r} 197 \\91\\70\end{tabular}\\
\hline
Clutter & \begin{tabular}{l} Yes \\No\end{tabular} & \begin{tabular}{r} 109 \\249\end{tabular} \\
\hline
Actuated DoFs & \begin{tabular}{l} Only internal \\ Internal and external \end{tabular} & \begin{tabular}{r} 162 \\ 196\end{tabular} \\
\hline
Interaction Wrenches & \begin{tabular}{l} Yes\\ No\end{tabular} & \begin{tabular}{r} 78\\ 280\end{tabular}
\end{tabular}
\end{table}

\subsection{Utilities}
We provide Python scripts and a ROS package on the website \href{https://tu-rbo.github.io/articulated-objects/}{https://tu-rbo.github.io/articulated-objects/} to facilitate the download and visualization of the data.
\begin{itemize}
\item The download script (\texttt{\footnotesize rbo\_downloader.py}) fetches object models and interaction files. The user can also select groups of interactions fulfilling a certain property, e.g. all interactions with an object, or all interactions with wrench measurements.
\item The visualization script (\texttt{\footnotesize rbo\_visualizer.py}) displays the content of an interaction folder: RGB, depth images, wrenches and/or joint states.
\item ROS package: Additionally to the interaction sequences in the file format described above, we also provide all data in the form of a \cite{rosbag}. We provide a ROS package including scripts to visualize the data in this format.
\end{itemize}


\section{Data Acquisition}
\label{s:data_acq}

\subsection{Visual Data and 6-D Body Poses}
The main goal of our dataset is to evaluate and develop algorithms based on visual data (RGB or RGB-D) with/without interaction wrenches for the perception of articulated objects. For this goal, it is crucial to register accurately the visual information and the ground truth provided by the motion capture system.
We first calibrate the intrinsic parameters of the RGB-D sensor. We use a checkerboard of known dimensions and take pictures at different poses of the checkerboard with respect to the camera with both the RGB and the infrared camera of the RGB-D sensor. We use an OpenCV-based camera calibration tool to estimate the internal parameters of the cameras (focal length, center point, and distortion parameters of the Plumb model) by detecting corner points on the checkerboard and estimating the parameters that minimize the squared error of the reprojection of the points from a separate multi-view PnP procedure per camera.
We then rectify the color and infrared images and estimate the 6-dimensional (6-D) transformation between the RGB and the infrared camera of the RGB-D sensor from a multi-view PnP procedure between the cameras.

We calibrate the extrinsic parameters of the RGB-D sensor with respect to the set of motion capture markers attached to it (see Fig.~\ref{fig:systemsetup}). We attached infrared markers to the corners of the checkerboard and use the motion capture system to detect their 3-D location. The points are projected on the color image based on the currently estimated transformation between the sensor and the marker set. We minimize the error of the projection of the point markers on the color image at different locations of the checkerboard.
After the calibration procedure, the point clouds recorded from the RGB-D sensor are registered with respect to the motion capture readings.

To acquire 6-D pose measurements of the articulated object from the motion capture system we attach marker sets to each of the links and place them inside the tracking volume.
The motion capture system estimates the 3-D location of the infrared fiducial markers on the scene with submillimeter accuracy and generates a 6-D pose measurement based on the predefined model of the arrangement of the markers within each marker set. While a minimum of three infrared markers defines a marker set we use at least five markers per set to improve the accuracy of the 6-D pose measurements and the robustness of the system against occlusions.

\subsection{Kinematic Properties}
\begin{figure}[t]
  \centering
  \includegraphics[width=.5\linewidth]{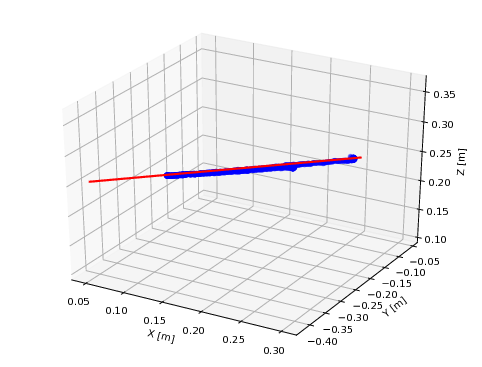}\includegraphics[width=.5\linewidth]{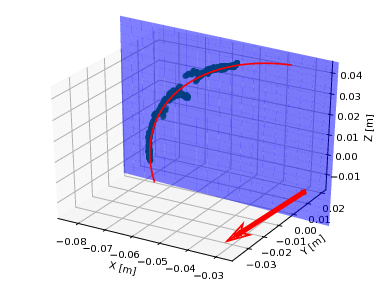}
  \caption{Estimating the joint axis for a prismatic (\textit{left}) and revolute joint (\textit{right}): The blue dots show the position of the moving body, the red lines are the fitted axes. For the revolute joint (\textit{right}) the orientation of the axis is obtained by fitting a plane (blue), while its position is based on a circle fit (red) within that plane.}
  \label{fig:fitting}
\end{figure}

We use the 6-D rigid body poses to compute the ground truth of the joint parameters in an offline batch procedure.
We estimate the axis of a prismatic joint by fitting a line to the time-varying positions of the child body with respect to the parent body in a least-squares sense~(Fig.~\ref{fig:fitting}, left).
The computation of the kinematic state of a prismatic joint requires to define an origin.
Without loss of generality we use the first pose of the child body with respect to the parent body as the origin.
We calculate the prismatic joint state as the distance between the child body and this point along the fitted line.

For revolute joints, we estimate the orientation of the joint axis as a unit vector, and its position as a 3-D point.
We obtain the plane that best fits the positions of the child body with respect to the parent body during the interaction by minimizing the squared distance of the points to the plane.
The plane's normal corresponds to the orientation of the revolute axis~(Fig.~\ref{fig:fitting}, right).
We then project all the points to the plane and estimate a circle using a least-squares fit with respect to the projected points.
The circle's center indicates the position of the revolute axis.
Without loss of generality we use the radius connecting the projection of the first pose of the child body with respect to the parent body as origin.
We calculate the configuration of the revolute joint as the angle of the arc between the child body and this point along the circle defined by joint axis position and orientation.

\subsection{Interaction Wrenches}
For each object we provide five interactions with measurements of the interaction wrench. In these interactions the humans actuate the articulated object with a tool attached to a force/torque (F/T) sensor~(Figure~\ref{fig:systemsetup}).
The motion capture system measures the pose of the interaction tool and we provide it as part of the interaction data.
We also provide a three-dimensional textured model of the tool with the sensor. The wrenches measured by the sensor and included in the dataset are raw values with bias. We measured the bias in the measurements with a calibration procedure where we align in turns one of the main axis of the F/T sensor to the vertical direction and collect the sensor readings. The result of the calibration is the following wrench bias vector: 
\begin{equation*}
w_{b} = \begin{pmatrix}f_b \\
 \tau_b\end{pmatrix} = 
\begin{pmatrix}(\SI{-0.927}{\N},\SI{1.122}{\N}, \SI{1.332}{\N})\\
(\SI{0.104}{\N\m}, \SI{0.027}{\N\m}, \SI{-0.033}{\N\m})
\end{pmatrix}
\end{equation*}
In order to subtract the effect of the tool from the wrench readings we measured the following dynamic properties of the elements attached to the F/T sensor
\begin{itemize}
\item Mass: \SI{163,0}{\g}
\item Center of Mass: (\SI{0}{\cm}, \SI{0}{\cm}, \SI{-1,5}{\cm})
\end{itemize}
and the transformation between the frame tracked by the motion capture system and the measurements frame depicted in Fig.~\ref{fig:systemsetup}:
\begin{equation*}
T_{ft\_mocap}^{ft\_meas} = 
\begin{pmatrix}
0.991 &  0.052 & -0.123 &  0.019 \\
-0.060 & -0.650 & -0.757 & -0.008 \\
-0.120 &  0.758 & -0.642 & -0.001 \\
0.0 &  0.0 &  0.0 & 1.0
\end{pmatrix}
\end{equation*}

%


\subsection{Link Geometries}
We use three alternative methods to generate three-dimensional triangle meshes of the links of the articulated models.
For small-scale objects, we use a system based on structured light~\citep{shaperec2}.
It projects a known light pattern onto the object to generate 3-D information.
The scanner acquires partial 3-D models from 24~different view points using a rotating plate and integrates them into a colored triangle mesh.
Before the scan we perform an initial calibration procedure to segment background from foreground.
For large-scale textured objects we use the software~\cite{shaperec}, which reconstructs a high definition 3-D mesh by applying a multi-view geometric algorithm on overlapping color photos of the object.
We take $\approx 25$~photos per object with a Casio Exilim EX-FC100 (resolution: 9~megapixel), located on a hemisphere centered around the object.
For large-scale textureless objects like the cabinet we generate meshes by hand using the 3-D creation suite Blender.
We post-process all models to fill holes, to remove parts of the surrounding environment, and to register them to the attached motion capture markers.

%

\begin{acks}
The authors would like to thank to Johannes Wortmann, Lorenz Vaitl, and Friedrich Meckel for their help in collecting the data, and Sebastian Koch for granting us access to and helping us with the 3-D scanner.
\end{acks}

\begin{funding}
We gratefully acknowledge the funding provided by the Alexander von Humboldt Foundation and the Federal Ministry of Education and Research (BMBF), by the European Commision (EC, SOMA, H2020-ICT-645599), and the German Research Foundation (DFG, Exploration Challenge, BR 2248/3-1).
\end{funding}


\bibliographystyle{SageH}
\bibliography{references}

\end{document}